\documentclass[runningheads]{llncs}
\usepackage{graphicx}
\usepackage{microtype}
\usepackage{booktabs} 

\usepackage{hyperref}

\usepackage{array}
\usepackage{amsmath}
\usepackage{amsbsy}
\usepackage{amssymb}
\usepackage{algorithm,algpseudocode}
\usepackage{float}
\usepackage{listings}
\usepackage{url}
\usepackage{tikz}
\usepackage{chngcntr}

\newcommand{\TODO}[1]{}

\newcommand{\Nexpert}{N}
\newcommand{\oexpert}[1]{o_{#1}}
\newcommand{\opool}{o^*}
\newcommand{\pexpert}[2]{p_{#1} \left(#2\right)}
\newcommand{\ppool}[1]{p^* \left(#1\right)}
\newcommand{\jexpert}[2]{j_{#1} \left(#2\right)}
\newcommand{\jpool}[1]{j^* \left(#1\right)}
\newcommand{\BNexpert}[1]{\mathcal{B}_{#1}}
\newcommand{\BNpool}{\mathcal{B}^*}
\newcommand{\Mexpert}[1]{\mathcal{M}_{#1}}
\newcommand{\Mpool}{\mathcal{M}^*}
\newcommand{\Mpoolprime}{\mathcal{M}'}

\newcommand{\jarule}[1]{\texttt{JARule}\left( #1 \right)}

\newcommand{\setremove}{ \backslash }
\newcommand{\setunion}{ \cup }

\newcommand{\exvarset}[1]{\mathcal{U}_{#1}}
\newcommand{\envarset}[1]{\mathcal{V}_{#1}}
\newcommand{\structeqset}[1]{\mathcal{F}_{#1}}
\newcommand{\removalset}[1]{\mathcal{W}_{#1}}

\newcommand{\model}[1]{\mathcal{M}_{#1}}
\newcommand{\modeldiagram}[1]{\mathcal{G}\left({#1}\right)}
\newcommand{\modeldef}{\left( \exvarset{}, \envarset{}, \structeqset{} \right)}
\newcommand{\probmodel}[1]{\mathcal{M}_{#1}}
\newcommand{\probmodeldistr}{P\left(U\right)}
\newcommand{\probmodeldef}{\left( \exvarset{}, \envarset{}, \structeqset{}, \probmodeldistr \right)}

\newcommand{\parentVset}[1]{\mathcal{V}_{pa_{#1}}}
\newcommand{\parentUset}[1]{\mathcal{U}_{pa_{#1}}}
\newcommand{\parentV}[1]{v_{pa_{#1}}}
\newcommand{\parentU}[1]{u_{pa_{#1}}}
\newcommand{\descendent}[2]{Desc_{#2}\left(#1\right)}

\newcommand{\context}[1]{\overrightarrow{#1}}

\newcommand{\vertexset}[1]{\mathsf{V}_{#1}}
\newcommand{\edgeset}[1]{\mathsf{E}_{#1}}
\newcommand{\edge}[2]{{#1 \rightarrow #2}}
\newcommand{\vspool}[1]{\mathsf{V}_{#1}^*}
\newcommand{\espool}[1]{\mathsf{E}_{#1}^*}

\newcommand{\predY}{\hat{Y}}
\newcommand{\predYfunction}{f_{\predY}}
\newcommand{\predYpool}{\hat{Y}^*}
\newcommand{\protectedattribset}[1]{\mathcal{A}}
\newcommand{\featureset}[1]{\mathcal{X}}
\newcommand{\fairset}[1]{\mathcal{Z}}
\newcommand{\unfairset}[1]{\mathcal{\bar{Z}}}

\begin{document}
\title{Counterfactually Fair Prediction Using Multiple Causal Models}
%
%
\author{Fabio Massimo Zennaro\orcidID{0000-0003-0195-8301} \and
Magdalena Ivanovska\orcidID{0000-0002-3916-3486}}
\authorrunning{F.M. Zennaro, M. Ivanovska}
%
\institute{Department of Informatics, University of Oslo, \\
	PO Box 1080 Blindern, 0316 Oslo, Norway \\
	\email{fabiomz@ifi.uio.no}\\
	\email{magdalei@ifi.uio.no} }
\maketitle              
\begin{abstract}
In this paper we study the problem of making predictions using multiple \emph{structural casual models} defined by different agents, under the constraint that the prediction satisfies the criterion of \emph{counterfactual fairness}. Relying on the frameworks of causality, fairness  and opinion pooling, we build upon and extend previous work focusing on the qualitative aggregation of causal Bayesian networks and causal models. In order to complement previous qualitative results, we devise a method based on Monte Carlo simulations. This method enables a decision-maker to aggregate the outputs of the causal models provided by different experts while guaranteeing the counterfactual fairness of the result. We demonstrate our approach on a simple, yet illustrative, toy case study.  

\end{abstract}

\section{Introduction}

In this paper we analyze the problem of integrating together the information provided in the form of multiple, potentially-unfair, predictive \emph{structural causal models} in order to generate predictions that are \emph{counterfactually fair}. 

This work is rooted in two main fields of research: \emph{causality} and \emph{fairness}. 
Causality deals with the definition and the study of causal relationships; structural causal models, in particular, are versatile and theoretically-grounded models that allow us to express causal relations and to study these relationships via interventions and counterfactual reasoning \cite{pearl2009causality}. 
Fairness is a research topic interested in evaluating if and how prediction systems deployed in sensitive scenarios may be guaranteed to support fair decisions; counterfactual fairness, in particular, is a concept of fairness developed in relation to causal models \cite{kusner2017counterfactual}.
The use of causal models in societally-sensitive contexts has been advocated by several researchers on the ground that the additional structure of these models and the possibility of evaluating the effect of interventions would allow for deeper understanding and control in critical situations \cite{barabas2017interventions}.
 
So far, little research has addressed the problem of aggregating multiple causal models. With no reference to fairness, \cite{bradley2014aggregating} studied a method to aggregate causal Bayesian networks, while \cite{alrajeh2018combining} introduced a notion of compatibility to analyze under which conditions causal models may be combined.
Taking fairness into account, \cite{russell2017worlds} proposed to tackle the problem of integrating multiple causal models as an optimization problem under the constraint of an $\epsilon$-relaxation of fairness.



In this paper we offer a complete solution for the problem of generating counterfactually-fair predictions given a set of causal models. Differently from \cite{bradley2014aggregating}, we focus our study on structural causal models instead of causal Bayesian networks, as the latter ones can encode causal relationships but do not support counterfactual reasoning \cite{pearl2009causality}; similarly to \cite{bradley2014aggregating}, though, we opt for a two-stage approach, made up of a qualitative stage, in which we work out the topology of an aggregated counterfactually-fair model, and a quantitative step, in which we use this topology to generate counterfactually-fair results out of individual causal models. The qualitative stage builds upon our previous work on this same topic in \cite{zennaro2018pooling}, and relies on the framework for judgment aggregation \cite{grossi2014judgment} and the work on pooling of causal Bayesian networks \cite{bradley2014aggregating}. The quantitative stage relies on Monte Carlo simulations \cite{mackay2003information} and, again, on opinion pooling \cite{dietrich2016probabilistic}.

The rest of the paper is organized as follows: Section \ref{sec:Background} reviews basic concepts in the research areas considered; Section \ref{sec:Aggregation} provides the formalization of our problem and our contribution; Section \ref{sec:Conclusion} draws conclusions on this work and indicates future avenues of research.

\section{Background \label{sec:Background}}

This section reviews basic notions used to define the problem of generating predictions out of multiple causal models under fairness: Section \ref{ssec:Causality} recalls the primary definitions in the study of causality; Section \ref{ssec:Fairness} discusses the notion of fairness in machine learning; Section \ref{ssec:OpPooling} offers a formalization of the problem of opinion pooling.

\subsection{Causality \label{ssec:Causality}}
Following the formalism in \cite{pearl2009causality}, we provide the basic definitions for working with causality.  

\paragraph{Causal Model.}
A \emph{(structural) causal model} $\model{}$ is a triple $\modeldef$ where:
\begin{itemize}
	\item $\exvarset{}$ is a set of exogenous variables $\left\lbrace U_1, U_2, ..., U_m \right\rbrace$ representing background factors that are not affected by other variables in the model;
	\item $\envarset{}$ is a set of endogenous variables $\left\lbrace V_1, V_2, ..., V_n \right\rbrace$ representing factors that are determined by other exogenous or endogenous variables in the model;
	\item $\structeqset{}$ is a set of functions $\left\lbrace f_1, f_2, ..., f_n \right\rbrace$, one for each variable $V_i$, such that the value $v_i$ is determined by the \emph{structural equation}: 
	$$v_i = f_i \left( \parentV{i}, \parentU{i} \right),$$ 
	where $\parentV{i}$ are the values assumed by the variables in the set $\parentVset{i} \subseteq \envarset{}\setremove \left\lbrace V_i \right\rbrace$ and $\parentU{i}$ are the values assumed by the variables in the set $\parentUset{i} \subseteq \exvarset{}$; that is, for each endogenous variable, there is a set of (parent) endogenous and a set of (parent) exogenous variables that determine its value through the corresponding structural equations.
\end{itemize}

\paragraph{Causal Diagram.}
The \emph{causal diagram} $\modeldiagram{\model{}}$ associated with the causal model $\model{}$ is the directed graph $\left(\vertexset{},\edgeset{}\right)$ where:
\begin{itemize}
	\item $\vertexset{}$ is the set of vertices representing the variables $\exvarset{} \setunion \envarset{}$ in $\model{}$;
	\item $\edgeset{}$ is the set of edges determined by the structural equations in $\model{}$; edges are coming to each endogenous node $V_i$ from each of its \emph{parent nodes} $\parentVset{i} \setunion \parentUset{i}$; we denote $\edge{V_j}{V_i}$ the edge going from $V_j$ to $V_i$. 
\end{itemize}
Assuming the acyclicity of causality, we will take that a causal model $\model{}$ entails an acyclic causal diagram $\modeldiagram{\model{}}$ represented as a \emph{directed acyclic graph} (DAG).

\paragraph{Context.}
Given a causal model $\model{} = \modeldef$, we define a \emph{context} $\context{u}=(u_1,u_2,\dots,u_m)$ as a specific instantiation of the exogenous variables, $U_1 = u_1, U_2 = u_2, \dots, U_m = u_m$.
Given an endogenous variable $V_i$, we will use the shorthand notation $V_i\left(\context{u}\right)$ to denote the value of the variable $V_i$ under the context $\context{u}$. This value is obtained by propagating the context $\context{u}$ through the causal diagram according to the structural equations.

\paragraph{Intervention.}
Given a causal model $\model{} = \modeldef$, we define \emph{intervention} $do(V_i = \bar{v})$ as the substitution of the structural equation $v_i=f_i \left( \parentV{i}, \parentU{i} \right)$ in the model $\model{}$ with the equation $v_i=\bar{v}$.
Given two endogenous variables $X$ and $Y$, we will use the shorthand notation $Y_{X\leftarrow x}$ to denote the value of the variable $Y$ under the intervention $do(X = x)$.\\
Notice that, from the point of view of the causal diagram, performing the intervention $do(X = x)$ is equivalent to setting the value of the variable $X$ to $x$ and removing all the incoming edges $\edge{\cdot}{X}$ in $X$.

\paragraph{Counterfactual.} 
Given a causal model $\model{} = \modeldef$, the context $\context{u}$, two endogenous variables $X$ and $Y$, and the intervention $do(X = x)$, a \emph{counterfactual} is the value of the expression $Y_{X\leftarrow x}(\context{u})$.\\
Note that, under the given context $\context{u}$, the variable $Y$ takes the value $Y(\context{u})$. Instead, the counterfactual  $Y_{X\leftarrow x}(\context{u})$ represents the value that $Y$ would have taken in the context $\context{u}$ had the value of $X$ been $x$.

\paragraph{Probabilistic Causal Model.}
A \emph{probabilistic causal model} $\probmodel{}$ is a tuple $\probmodeldef$ where:
\begin{itemize}
\item $\modeldef$ is a causal model;
\item $\probmodeldistr$ is a probability distribution over the exogenous variables. The probability distribution $\probmodeldistr$, combined with the dependence of each endogenous variable $V_i$ on the exogenous variables, as specified in the structural equation for $v_i$, allows us to define a probability distribution $P(V)$ over the endogenous variables as: $P(V=v)=\sum_{\{ \context{u} \vert V = v \}} P(U= \context{u})$.
\end{itemize} 
Notice that we overload the notation $\probmodel{}$ to denote both (generic) causal models and probabilistic causal models; the context will allow the reader to distinguish between them.

\subsection{Fairness \label{ssec:Fairness}}
Following the work of \cite{kusner2017counterfactual}, we review the topic of fairness, with a particular emphasis on counterfactual fairness for predictive models.

\paragraph{Fairness and Learned Models.}
Black-box machine learning systems deployed in sensitive contexts (e.g.: police enforcement or educational grants allocation) and trained on historical real-world data have the potential of perpetuating, or even introducing \cite{kusner2017counterfactual}, socially or morally unacceptable discriminatory biases (for a survey, see, for instance, \cite{zliobaite2015survey}). The study of \emph{fairness} is concerned with the definition of new metrics to assess and guarantee the social fairness of a predictive decision system.

\paragraph{Fairness of a Predictor.}
A predictive model can be represented as a (potentially probabilistic) function of the form $\predY = f(Z)$, where $\predY$ is a \emph{predictor} and $Z$ is a vector of covariates. An observational approach to fairness states that the set of covariates can be partitioned in a set of \emph{protected attributes} $\protectedattribset{}$, representing discriminatory elements of information, and a set of \emph{features} $\featureset{}$, carrying no sensitive information.
The predictive model can then be redefined as $\predY = f(A,X)$ and the fairness problem is expressed as the problem of learning a predictor $\predY$ that does not discriminate with respect to the protected attributes $\protectedattribset{}$. 
Given the complexities of social reality and disagreement over what constitutes a fair policy, different measures of fairness may be adopted to rule out discrimination (e.g.: counterfactual fairness or fairness through unawareness); for a more thorough review of different types of fairness and their limitations, see \cite{gajane2017formalizing} and \cite{kusner2017counterfactual}.

\paragraph{Fairness Over Causal Models.}
Given a probabilistic causal model $\probmodeldef$ fairness may be evaluated following an observational approach. Let us take $\predY$ to be an endogenous variable whose structural equation provides the predictive function $\predYfunction$; let us also partition the remaining variables $\exvarset{} \cup \envarset{} \setminus \{\predY\}$ into a set of protected attributes $\protectedattribset{}$ and a set of features $\featureset{}$. Then we can evaluate the fairness of the predictor $\predY$ with respect to the discriminatory attributes $\protectedattribset{}$.

\paragraph{Counterfactual Fairness.}
Given a probabilistic causal model $\probmodel{} = \probmodeldef$, a predictor $\predY$, and a partition of the variables $\exvarset{} \cup \envarset{} \setminus \{\predY\}$ into $\left(\protectedattribset{}, \featureset{} \right)$, the predictor $\predYfunction$ is \emph{counterfactually fair} if, for every context $\context{u}$, $$ P\left( \predY_{A\leftarrow a} (\context{u}) \vert X=x, A=a \right) =  
P\left( \predY_{A\leftarrow a'} (\context{u}) \vert X=x, A=a \right),$$ for all values $y$ of the predictor, for all values $a'$ in the domain of $A$, and for all $x$ in the domain of $X$ \cite{kusner2017counterfactual}.\\
In other words, the predictor $\predY$ is counterfactually fair if, under all the contexts, the prediction on $\predY$ given the observation of the protected attributes $A=a$ and the features $X=x$ would not change if we were to intervene $do(A=a')$ to force the value of the protected attributes $A$ to $a'$, for all the possible values that the protected attributes can assume.\\
Denoting $\descendent{\protectedattribset{}}{\model{}}$ the descendants of the nodes in $\protectedattribset{}$ in the model $\model{}$, an immediate property follows from the definition of counterfactual fairness:
\begin{lemma}\label{lemma}
(Lemma 1 in \cite{kusner2017counterfactual}) Given a probabilistic causal model $\probmodeldef$, a predictor $\predY$ and a partition of the variables into $\left( \protectedattribset{}, \featureset{} \right)$, the predictor $\predY$ is counterfactually fair if $\predYfunction$ is a function depending only on variables that are not in $\descendent{\protectedattribset{}}{\model{}}$.
\end{lemma}

\subsection{Opinion Pooling \label{ssec:OpPooling}}
Following the study of \cite{dietrich2016probabilistic}, we introduce the framework for opinion pooling.

\paragraph{Opinion Pooling.} Assume there are $\Nexpert$ experts, each one expressing his/her opinion $\oexpert{i}$, $1 \leq i \leq \Nexpert$. The problem of \emph{pooling} (or \emph{aggregating}) the opinions $\oexpert{i}$ consists in finding a single pooled opinion $\opool$ representing the collective opinion that best represents the individual opinions in the given context.

\paragraph{Probabilistic Opinion Pooling.}

Given opinions in the form of probability distributions $\pexpert{i}{x}$, $1 \leq i \leq \Nexpert$, defined over the same domain, \emph{probabilistic opinion pooling} is concerned with finding a single pooled probability distribution $\ppool{x}= F\left(p_1,\ldots,p_n\right)(x)$, where $F$ is a functional mapping a tuple of pdfs to a single probability distribution \cite{dietrich2016probabilistic}.\\
Now, given a set of probabilistic opinions $\pexpert{i}{x}$, different functionals $F$ may be chosen to perform opinion pooling, either using a principled approach such as an axiomatic approach based on the definition of a set of desired properties \cite{dietrich2016probabilistic}, or using standard statistical operators, such as arithmetic averaging or geometric averaging. 

\paragraph{Judgment Aggregation.}
Given opinions in the form of a set of Boolean functions $\jexpert{i}{x}$, \emph{judgment aggregation} is concerned with finding a single Boolean function $\jpool{x}= F(j_1,\ldots,j_n)(x)$, where $F$ is a functional mapping a tuple of Boolean functions to a single Boolean function \cite{grossi2014judgment}.\\
As in the case of probabilistic opinion pooling, given a set of judgments $\jexpert{i}{x}$, different functions $F$ may be chosen to perform judgment aggregation, such as majority voting or intersection  \cite{bradley2014aggregating}.



\paragraph{Aggregation of Causal Bayesian Networks.}
Given opinions expressed in the form of causal Bayesian networks\footnote{
A Bayesian network (BN) \cite{pearl2014probabilistic}
is a structured representations of the joint probability distribution of a set of variables in the form of a directed acyclic graph with associated conditional probability distributions. A causal BN is a BN where all the edges represent causal relations between variables.} (CBN) $\BNexpert{i}$, aggregation of CBNs is concerned with defining a single pooled CBN $\BNpool$.\\	
A seminal study in aggregating CBNs is offered by 	 \cite{bradley2014aggregating}. They suggest a \emph{two-stage approach} to the problem of aggregating $\Nexpert$ causal Bayesian networks $\BNexpert{i}$. In the first qualitative stage, they determine the graph of the pooled CBN reducing the problem to a judgment aggregation over the edges in the individual CBNs; namely, for every two variables $X$ and $Y$, the presence of an edge from node $X$ to node $Y$ in the model $\BNexpert{i}$ is represented as the $i$-th expert casting the judgment $\jexpert{i}{\edge{X}{Y}}=1$, and the absence of it as the judgment $\jexpert{i}{\edge{X}{Y}}=0$; the problem of defining the pooled graph is then reduced to a judgment aggregation problem over the judgments $\jexpert{i}{}$.
In the second quantitative step, they derive the conditional probability distributions for the pooled graph applying probabilistic opinion pooling to the corresponding conditional probability distributions in the individual CBNs.
A critical result in the study of \cite{bradley2014aggregating} is the translation of the classical impossibility theorem from judgment aggregation \cite{grossi2014judgment} into an impossibility theorem for the qualitative aggregation of CBNs:
\begin{theorem} \label{theorem}
	(Theorem 1 in \cite{bradley2014aggregating}) Given a set of CBNs defined over at least three variables, there is no judgment aggregation rule $f$ that satisfies all the following properties:
	\begin{itemize}
		\item \emph{Universal Domain}: the domain of $f$ includes all logically possible acyclic causal relations;
		\item \emph{Acyclicity}: the pooled graph produced by $f$ is guaranteed to be acyclic over the set of nodes;
		\item \emph{Unbiasedness}: given two variables $X$ and $Y$, the causal dependence of $X$ on $Y$ in the pooled graph rests only on whether $X$ is causally dependent on $Y$ in the individual graphs, and the aggregation rule is not biased towards a positive or negative outcome;
		\item \emph{Non-dictatorship}: the pooled graph produced by $f$ is not trivially equal to the graph provided by one of the experts.
	\end{itemize} 
\end{theorem} 
As a consequence of this theorem, no unique aggregation rule satisfying the above properties can be chosen for the pooling of causal judgments in the first step of the two-stage approach. A relaxation of these properties must be decided depending on the scenario at hand.

\section{Aggregation of Causal Models Under Fairness \label{sec:Aggregation}}

This section analyzes how probabilistic causal models can be aggregated under fairness: Section \ref{ssec:ProblemFormal} provides a formalization of our problem; Section \ref{ssec:QualitativeStep} discusses how to define the topology of a counterfactually-fair causal graph by pooling together the graphs of different probabilistic structural causal models; Section \ref{ssec:QuantitativeStep} explains how to evaluate a counterfactually-fair prediction out of the individual models using the pooled graph; finally, Section \ref{ssec:Illustration} offers an illustration of the use of our method on a toy case study.

\subsection{Problem Formalization \label{ssec:ProblemFormal}}

Let us consider the case in which $\Nexpert$ experts are required to provide a probabilistic causal model $\Mexpert{i} = \left( \exvarset{}, \envarset{}, \structeqset{i}, P_i(U) \right)$ representing a potentially socially-sensitive scenario.
For simplicity, we assume that the experts are provided with a fixed set of variables $\left(\exvarset{}, \envarset{}\right)$. The task of the experts can be summarized in two modeling phases: (i) a qualitative phase, in which they define the causal topology of the graph $\modeldiagram{\model{i}}$ (\emph{which} variables are causally influencing which other variables); and, (ii) a quantitative phase, specifying the probability distribution functions $P_i(U)$ (\emph{how} the stochastic behavior of the exogenous variables is modeled) and the structural equations $\structeqset{i}$ (\emph{how} each endogenous variable is causally influenced by its parents variables).

Critically, we are not requesting the experts to provide fair models. Individual experts may not be aware of specific discrimination issues, they may have different understandings of fairness or, simply, they may not have the technical competence to formally evaluate or guarantee fairness. The task of defining which form of fairness is relevant, and enforcing it, is up to the final decision-maker only. 


The (potentially unfair) models $\Mexpert{i}$ defined by the experts are then provided to a decision-maker, who wants to exploit them to compute a single counterfactually-fair predictive output $\predYpool$. We assume the decision-maker to be knowledgeable of fairness implications and to be responsible for partitioning the exogenous and endogenous variables $\exvarset{} \cup \envarset{}$ into sensitive attributes $\protectedattribset{}$ and non-sensitive attributes $\featureset{}$.

In summary, our problem may be expressed as follows: given $\Nexpert$ (potentially counterfactually-unfair) probabilistic causal models $\Mexpert{i}$ defined on the same variables $\left(\exvarset{}, \envarset{}\right)$, and a partition of the variables into $\left( \protectedattribset{}, \featureset{} \right)$, can we define a pooling algorithm that allows us to construct an aggregated counterfactually-fair causal model $\Mpool = f\left(\Mexpert{i}\right)$ and compute a counterfactually-fair predictive output $\predYpool$?

Mirroring the modeling approach of the experts, we propose to solve this problem by adopting a two-stage approach. We reduce the task of the final decision-maker to the two following phases: (i) a qualitative phase, in which we compute the topology of a counterfactually-fair pooled graph $\modeldiagram{\Mpool}$ (\emph{which} nodes and causal links from the individual models can be retained under a requirement of counterfactual fairness); and, (ii) a quantitative phase, in which we provide a method to evaluate a probability distribution over the final predictor $\predYpool$ (\emph{how} is the final counterfactually-fair predictor computed).

\subsection{Qualitative Aggregation over the Graph \label{ssec:QualitativeStep}}

In the qualitative phase of our approach, we consider the different expert models $\Mexpert{i}$ and we focus on the problem of defining the topology of an aggregated graph $\modeldiagram{\Mpool}$ that guarantees counterfactual fairness.

We tackle this challenge following the solution proposed in \cite{bradley2014aggregating} to perform a qualitative aggregation of causal Bayesian networks (see Section \ref{ssec:OpPooling}): in each causal model $\Mexpert{i}$, we convert each edge in a binary judgment and we then perform judgment aggregation according to a chosen aggregation rule $\jarule{}$.
However, this solution presents two shortcomings: (i) it does not guarantee that the predictor $\predYpool$ in the aggregated probabilistic causal model $\Mpool$ will be counterfactually fair; and (ii) because of Theorem \ref{theorem}, we cannot apply this procedure without first choosing one of the properties in the theorem statement to sacrifice.
We solve the first problem by relying and enforcing the condition specified in Lemma \ref{lemma}; practically, we introduce in our algorithm a \emph{removal step}, in which we remove all the protected attributes and their descendants from the aggregated model $\Mpool$. This procedure satisfies by construction the condition in Lemma $\ref{lemma}$, and thus guarantees counterfactual fairness.
We address the second problem by arguing that the structure of the causal graph immediately suggests the possibility of dropping the property of unbiasedness, which requires that the presence of an edge in the pooled graph depends only on the presence of the same edge in each individual graph; we can relax this property, by making the presence of an edge in the final graph dependent also on an ordering of the edges. In our specific case, we can easily introduce an ordering of the edges with respect to the predictor $\predY$, as a starting point. The ordering produced in this way may not be total, and we may still have to introduce another rule to break potential ties (e.g., random selection or an alphabetical criterion). 
We formalize this procedure in an additional algorithmic step, \emph{pooling step}, in which we order edges in relation to their distance from the predictor and then we perform judgment aggregation using the rule $\jarule{}$ so that, if the edge is selected for insertion in the pooled model $\modeldiagram{\Mpool}$, then it is added as long as acyclicity is not violated \cite{bradley2014aggregating}.

The two algorithmic steps defined above may be interchangeably combined. This gives rise to two alternative algorithms: a \emph{removal-pooling} algorithm (Algorithm \ref{alg:removal-pooling}), and a \emph{pooling-removal} algorithm (Algorithm \ref{alg:pooling-removal}). 
We provide the pseudo code for the two functions, $\mathsf{Removal}()$ and $\mathsf{Pooling}()$, in Algorithm \ref{alg:removal} and \ref{alg:pooling} in the appendix; also, for a detailed description and analysis of these functions and their outputs we refer the reader to \cite{zennaro2018pooling}.

At the end, both algorithms return the skeleton of a pooled causal model $\Mpool$ that is guaranteed to be counterfactually fair. However, the result, so far, contains only a qualitative description of the causal model: we determined the topology of the graph $\modeldiagram{\Mpool}$, but the structural equations of the pooled causal model $\Mpool$ are left undefined.

\begin{algorithm} 
	\caption{Removal-Pooling Algorithm}\label{alg:removal-pooling}
	\begin{algorithmic}[1]
		\State {\bfseries Input:} $\Nexpert$ graph models $\modeldiagram{\model{i}}$ over the variables $\{\exvarset{},\envarset{}\}$, a predictor $\predY \in \envarset{}$, a partitioning of the variables $\{\exvarset{},\envarset{}\} \setminus \{\predY\}$ into protected attributes $\protectedattribset{}$ and $\featureset{}$, a judgment aggregation rule $\jarule{}$  
		\State
		\State $\{\model{i}'\}_{i=1}^N = \mathsf{Removal}\left(\{\model{i}\}_{i=1}^N, \protectedattribset{}\right)$
		\State $\Mpool = \mathsf{Pooling}\left(\{\model{i}'\}_{i=1}^N, \jarule{}\right)$
		\State \Return $\Mpool$
	\end{algorithmic}
\end{algorithm}

\begin{algorithm} 
	\caption{Pooling-Removal Algorithm}\label{alg:pooling-removal}
	\begin{algorithmic}[1]
		\State {\bfseries Input:} $\Nexpert$ graph models $\modeldiagram{\model{i}}$ over the variables $\{\exvarset{},\envarset{}\}$, a predictor $\predY \in \envarset{}$, a partitioning of the variables $\{\exvarset{},\envarset{}\} \setminus \{\predY\}$ into protected attributes $\protectedattribset{}$ and $\featureset{}$, a judgment aggregation rule $\jarule{}$  
		\State
		\State $\Mpoolprime = \mathsf{Pooling}\left(\{\model{i}\}_{i=1}^N, \jarule{}\right)$
		\State $\Mpool = \mathsf{Removal}\left(\Mpoolprime, \protectedattribset{}\right)$
		\State \Return $\Mpool$
	\end{algorithmic}
\end{algorithm}

\subsection{Quantitative Aggregation over the Distribution of the Predictor \label{ssec:QuantitativeStep}}

In the quantitative phase of our approach, we study how we can use the topology of the pooled and counterfactually-fair causal graph $\modeldiagram{\Mpool}$ that we have generated in the first step to produce quantitative and counterfactually-fair outputs.

After the phase of qualitative aggregation, the individual models provided by the experts $\Mexpert{i}$ have been aggregated only with respect to their nodes and edges; the pooled graph $\Mpool$ encodes the topology of a counterfactually-fair causal model, but it lacks the definition of probability distributions over the exogenous nodes and structural equations over the endogenous nodes in order to be complete and usable.

Defining the probability distributions and the structural equations in the aggregated model $\Mpool$ by pooling together individual functions in each expert model $\Mexpert{i}$ is a particularly challenging task: different experts may provide substantially different functions, functions may be defined on different domains (since the same node may have different incoming edges in different expert graphs), and domains may have been changed in the aggregated model (since nodes may have been dropped in order to guarantee counterfactual fairness).
Therefore, instead of finding an explicit form for the probability distributions and the structural equations in the aggregated model $\Mpool$, we suggest to compute the distribution of the predictor $\predY_i$ in each expert model $\Mexpert{i}$ while \emph{integrating out} all the components that do not belong to the counterfactually-fair graph $\modeldiagram{\Mpool}$, and finally aggregate them to obtain the final counterfactually-fair predictor $\predYpool$.

More formally, let $\fairset{} \subseteq \exvarset{} \cup \envarset{}$ be the set of \emph{fair features} corresponding to nodes that are present in $\modeldiagram{\Mpool}$, and let $\unfairset{} \subseteq \exvarset{} \cup \envarset{}$ be the set of \emph{unfair features} corresponding to nodes that are not present in $\modeldiagram{\Mpool}$.
Now, if we are given an instance of fair features $Z=z$, we can compute the probability distribution of the predictor $\predY_i$ in each model $\Mexpert{i}$ by integrating out the unfair features:
\[ 
P(\predY_i \vert Z=z) = \int P(\predY_i \vert Z=z, \bar{Z}) d\bar{Z}.
\] 
In other words, we use the aggregated model $\modeldiagram{\Mpool}$ to identify in each expert model a countefactually fair sub-graph and to integrate out the contributions of the rest of the graph.
Practically, this operation of integration and estimation of the distribution of $P(\predY_i \vert Z=z)$ can be efficiently carried out using Monte Carlo sampling \cite{mackay2003information}.

The final result will then consist of a set of $\Nexpert$ individual pdfs $P(\predY_i \vert Z=z)$. In order to take a decision, these pdfs can be simply merged together using simple statistical operators (e.g.: by taking the mean of the expected values) or relying on standard opinion pooling operators \cite{dietrich2016probabilistic}.

\subsection{Illustration \label{ssec:Illustration}}

Here we give a simple illustration of the problem of causal model aggregation under counterfactual fairness, which we recover from \cite{zennaro2018pooling}.

\paragraph{Setup.}
In this example, we imagine that the head of a Computer Science department asked two professors, Alice and Bob, to design a predictive system to manage PhD selections. In particular, we imagine that Bob and Alice were required to define a causal model over the endogenous variables \emph{age} (Age), \emph{gender} (Gnd), \emph{MSc university department} (Dpt), \emph{MSc final mark} (Mrk), \emph{experience in the job market} (Job), \emph{quality of the cover letter} (Cvr), the relative exogenous variables (U$_\cdot$) and a predictor ($\predY$).  

The graphs of the two models provided by Alice $\modeldiagram{\model{A}}$ and Bob $\modeldiagram{\model{B}}$ are given in Figures \ref{fig:M_A} and \ref{fig:M_B}, respectively.
	
	\begin{figure}
	\begin{minipage}{.5\textwidth}	
	\begin{tikzpicture}[shorten >=1pt, auto, node distance=3cm, thick, scale=0.8, every node/.style={scale=0.8}]
	
	\tikzstyle{node_style} = [circle,draw=black]
	\node[] (Uage) at (4,2) {$U_{age}$};
	\node[] (Ujob) at (2,1) {$U_{job}$};
	\node[] (Ugnd) at (0,1) {$U_{gnd}$};
	\node[] (Udpt) at (0,-3) {$U_{dpt}$};
	\node[] (Umrk) at (2,-4) {$U_{mrk}$};
	\node[] (Ucvr) at (4,-4) {$U_{cvr}$};

	\node[node_style] (gnd) at (0,0) {Gnd};
	\node[node_style] (dpt) at (0,-2) {Dpt};
	\node[node_style] (job) at (2,0) {Job};
	\node[node_style] (mrk) at (2,-3) {Mrk};
	\node[node_style] (age) at (4,1) {Age};
	\node[node_style] (cvr) at (4,-3) {Cvr};
	\node[circle,draw=black,dashed] (y) at (6,-1) {$\predY$};
	
	\draw[->]  (Uage) edge (age);
	\draw[->]  (Ujob) edge (job);
	\draw[->]  (Ugnd) edge (gnd);
	\draw[->]  (Udpt) edge (dpt);
	\draw[->]  (Umrk) edge (mrk);
	\draw[->]  (Ucvr) edge (cvr);
	
	\draw[->]  (gnd) edge (job);
	\draw[->]  (gnd) edge (dpt);
	\draw[->]  (dpt) edge (mrk);
	\draw[->]  (age) edge (job);
	
	\draw[->]  (cvr) edge (y);
	\draw[->]  (job) edge (y);
	\draw[->]  (dpt) edge (y);
	\draw[->]  (mrk) edge (y);
	
	\end{tikzpicture}
	
	\caption{Graph $\modeldiagram{\model{A}}$. \label{fig:M_A}}
\end{minipage}
\begin{minipage}{.5\textwidth}	
	\begin{tikzpicture}[shorten >=1pt, auto, node distance=3cm, thick, scale=0.8, every node/.style={scale=0.8}]
	
	\tikzstyle{node_style} = [circle,draw=black]
	\node[] (Uage) at (4,2) {$U_{age}$};
	\node[] (Ujob) at (2,1) {$U_{job}$};
	\node[] (Ugnd) at (0,1) {$U_{gnd}$};
	\node[] (Udpt) at (0,-3) {$U_{dpt}$};
	\node[] (Umrk) at (2,-4) {$U_{mrk}$};
	\node[] (Ucvr) at (4,-4) {$U_{cvr}$};
	
	\node[node_style] (gnd) at (0,0) {Gnd};
	\node[node_style] (dpt) at (0,-2) {Dpt};
	\node[node_style] (job) at (2,0) {Job};
	\node[node_style] (mrk) at (2,-3) {Mrk};
	\node[node_style] (age) at (4,1) {Age};
	\node[node_style] (cvr) at (4,-3) {Cvr};
	\node[circle,draw=black,dashed] (y) at (6,-1) {$\predY$};
	
	\draw[->]  (Uage) edge (age);
	\draw[->]  (Ujob) edge (job);
	\draw[->]  (Ugnd) edge (gnd);
	\draw[->]  (Udpt) edge (dpt);
	\draw[->]  (Umrk) edge (mrk);
	\draw[->]  (Ucvr) edge (cvr);
	
	\draw[->]  (gnd) edge (job);
	\draw[->]  (dpt) edge (mrk);
	\draw[->]  (age) edge (job);
	
	\draw[->]  (cvr) edge (y);
	\draw[->]  (job) edge (y);
	\draw[->]  (dpt) edge (y);
	\draw[->]  (mrk) edge (y);
	\draw[->]  (age) edge (y);
	
	\end{tikzpicture}
	
	\caption{Graph $\modeldiagram{\model{B}}$. \label{fig:M_B}}
	
\end{minipage}
\end{figure}

Moreover, Alice and Bob came up with the following probability distributions over the exogenous nodes $\exvarset{}$:

\footnotesize
\begin{minipage}{.5\textwidth}
\begin{align*}
U_{age\_A} & \sim\textrm{Poisson}(\lambda=3)\\
U_{job\_A} & \sim\textrm{Bernoulli}(p=0.3)\\
U_{gnd\_A} & \sim\textrm{Bernoulli}(p=0.5)\\
U_{dpt\_A} & \sim\textrm{Categorical}([0.7,0.2,0.1])\\
U_{mrk\_A} & \sim\textrm{Beta}(\alpha=2,\beta=2)\\
U_{cvr\_A} & \sim\textrm{Beta}(\alpha=2,\beta=5)
\end{align*}
\end{minipage}
\begin{minipage}{.5\textwidth}
\begin{align*}
U_{age\_B} & \sim\textrm{Poisson}(\lambda=4)\\
U_{job\_B} & \sim\textrm{Bernoulli}(p=0.2)\\
U_{gnd\_B} & \sim\textrm{Bernoulli}(p=0.5)\\
U_{dpt\_B} & \sim\textrm{Categorical}([0.7,0.15,0.1,0.05])\\
U_{mrk\_B} & \sim\textrm{Beta}(\alpha=2,\beta=2)\\
U_{cvr\_B} & \sim\textrm{Beta}(\alpha=2,\beta=5)
\end{align*}
\end{minipage}
\normalsize

\vspace{10pt}
and they defined the structural equations for the endogenous nodes $\envarset{}$:
\footnotesize
\begin{align*}
V_{age\_A} & =20+U_{age\_A}\\
V_{gnd\_A} & =U_{gnd\_A}\\
V_{job\_A} & =U_{job\_A}+\frac{V_{gnd\_A}}{2}+\frac{V_{age\_A}}{100}\\
V_{dpt\_A} & =\textrm{if \ensuremath{\left(V_{gnd\_A}=1\right)} then \ensuremath{0} else \ensuremath{\frac{U_{dpt\_A}}{10}}}\\
V_{mrk\_A} & =\textrm{if \ensuremath{\left(V_{dpt\_A}=0\right)} then \ensuremath{U_{mrk\_A}+0.1} else \ensuremath{U_{mrk\_A}-0.1}}\\
V_{cvr\_A} & =U_{cvr\_A}+0.2\\
\hat{Y}_{A} & =V_{job\_A}+V_{dpt\_A}+V_{mrk\_A}+V_{cvr\_A}
\end{align*}
\begin{align*}
V_{age\_B} & =19+U_{age\_B}\\
V_{gnd\_B} & =U_{gnd\_B}\\
V_{job\_B} & =U_{job\_B}+\frac{V_{age\_B}}{100}+\left[\textrm{if \ensuremath{\left(V_{gnd\_B}=1\right)} then \ensuremath{0.5} else \ensuremath{0}}\right]\\
V_{dpt\_B} & =\textrm{\ensuremath{\frac{U_{dpt\_B}}{10}}}\\
V_{mrk\_B} & =\textrm{if \ensuremath{\left(V_{dpt\_B}=0\right)} then \ensuremath{U_{mrk\_B}+0.1} else \ensuremath{U_{mrk\_B}}}\\
V_{cvr\_B} & =U_{cvr\_B}+0.1\\
\hat{Y}_{A} & =\frac{V_{job\_A}}{100}+V_{job\_A}+V_{dpt\_A}+V_{mrk\_A}+V_{cvr\_A}
\end{align*}
\normalsize
Notice that these probability distributions and structural equations are pure examples, and do not have any deep meaningful relation with the scenario at hand; they were chosen mainly to illustrate the use of a variety of distributions and functions, and to output as a predictor $\predY$ a score that can be used for decision-making. In reality, such functions would be determined via machine learning methods or carefully defined by a modeler.

Notice that the models and the structural equations were defined by Alice and Bob with no explicit concern about any form of fairness. Now, however, the head of the department wants to aggregate these models in a way that guarantees counterfactual fairness with respect to the gender of the PhD candidate.

\paragraph{Qualitative Aggregation.}
As a first step, the head of the department assumes all the exogenous variables to be non-sensitive and partitions the endogenous ones into protected attributes $\protectedattribset{}=\{\textnormal{Gnd}\}$ and features $\featureset{}=\{\textnormal{Age},\textnormal{Dpt},\textnormal{Mrk},\textnormal{Job},\textnormal{Cvr}\}$, and then chooses as a judgment aggregation rule $\jarule{}$ the \emph{strict majority rule}.

She then decides to apply the \emph{pooling-removal} algorithm to the models $\Mexpert{A}$ and $\Mexpert{B}$. A detailed explanation of the application of this algorithm is available in \cite{zennaro2018pooling}.
The resulting pooled counterfactually-fair model $\Mpool$ is illustrated in Figure \ref{fig:Mstar}.
\begin{figure}
	\centering
	
	\begin{tikzpicture}[shorten >=1pt, auto, node distance=3cm, thick, scale=0.8, every node/.style={scale=0.8}]
	
	\tikzstyle{node_style} = [circle,draw=black]
	\node[] (Udpt) at (4,-3) {$U_{dpt}$};
	\node[] (Umrk) at (5.5,-4) {$U_{dpt}$};
	\node[] (Ucvr) at (7,-4) {$U_{dpt}$};
	
	\node[node_style] (cvr2) at (7,-3) {Cvr};
	\node[circle,draw=black,dashed] (y2) at (8,-1) {$\predY$};	
	\node[node_style] (dpt2) at (4,-2) {Dpt};
	\node[node_style] (mrk2) at (5.5,-3) {Mrk};
	
	\draw[->]  (Udpt) edge (dpt2);
	\draw[->]  (Umrk) edge (mrk2);
	\draw[->]  (Ucvr) edge (cvr2);
	
	\draw[->]  (cvr2) edge (y2);
	\draw[->]  (dpt2) edge (y2);
	\draw[->]  (mrk2) edge (y2);
	\draw[->]  (dpt2) edge (mrk2);
	
	\end{tikzpicture}
	
	\caption{Graph of the pooled counterfactually-fair model $\modeldiagram{\Mpool}$ after applying the \emph{pooling-removal} algorithm. \label{fig:Mstar}}
\end{figure}

\paragraph{Quantitative Aggregation.}

At this point, the head of the department can use the individual expert models $\Mexpert{A}$ and $\Mexpert{B}$, and the aggregated counterfactually-fair model $\modeldiagram{\Mpool}$ to compute predictive scores for the PhD applicants.

Suppose, for instance, that the following candidates were to submit their application:
\begin{align*}
\textrm{App}_{1} & =\left\{ \textrm{Age=22; Gnd=F; Dpt=Computer Science; Mrk=0.8; Job=True; Cvr=0.4}\right\} \\
\textrm{App}_{2} & =\left\{ \textrm{Age=22; Gnd=M; Dpt=Computer Science; Mrk=0.8; Job=True; Cvr=0.4}\right\} 
\end{align*}
From a formal point of view, we may imagine the second candidate as the result of the intervention $do(\textnormal{Gnd}=M)$ on the first candidate, thus forcing the gender to male.

Now, for the sake of illustration, we implemented the models and the candidates using Edward \cite{tran2016edward}, a Python library for probabilistic modeling, and we made the code available online\footnote{\url{https://github.com/FMZennaro/Fair-Pooling-Causal-Models}}. Whenever using Monte Carlo sampling, we collected $10^5$ samples. 

The models provided by Alice and Bob obviously define two different predictors $\predY_A$ and $\predY_B$ with two dissimilar probability distributions (see Figure \ref{fig:P_Y} in the appendix for an estimation of these pdfs). If the head of the department were to feed the data about the candidates to the two models, she would receive different and unfair results:
\[
\begin{array}{ccc}
\predY_A\left(\textrm{App}_{1}\right)=4.720 & \,\,\,\,\,\, & \predY_B\left(\textrm{App}_{1}\right)=3.340\\
\predY_A\left(\textrm{App}_{2}\right)=2.720 &  & \predY_B\left(\textrm{App}_{2}\right)=2.840
\end{array}
\]
Indeed, these results show both a legitimate disagreement between Alice and Bob on how they score individual candidates, but they also show a troubling internal disagreement in that both experts assign different scores to the two applicants when the only difference between them is their gender. Since the head of the department considers gender a protected attribute, this result is deemed unfair. More formally, if the second candidate were to be seen as an intervention we would have:
\[ 
P\left( \predY_{\textnormal{Gnd}\leftarrow F} (\context{u}) \vert X=x, A=a \right) \neq  
P\left( \predY_{\textnormal{Gnd}\leftarrow M} (\context{u}) \vert X=x, A=a \right),
\]
thus denying counterfactual fairness.

To tackle the problem, the head of the department decides to evaluate the predictive score for the candidates by computing the distribution of the predictor $\predY_i$ given only the fair features $\fairset{}$ from the pooled counterfactually-fair model $\Mpool$:
\[ 
P(\predY_A \vert Z=z) = \int_{Age,Gnd,Job} P(\predY_A \vert  Dpt=\textnormal{CS}, Mrk=0.8, Cvr=0.4)
\]
\[ 
P(\predY_B \vert Z=z) = \int_{Age,Gnd,Job} P(\predY_B \vert  Dpt=\textnormal{CS}, Mrk=0.8, Cvr=0.4)
\]
This step does not provide a scalar output as in the previous evaluation, but it defines two probability distributions $P(\predY_A \vert Z=z)$ and $P(\predY_B \vert Z=z)$ (see Figure \ref{fig:P_Y_given_Z} in the appendix for an estimation of these pdfs). These two pdfs can now be pooled together for final decision making.
After deciding to compute the average of the expected value of the pdfs\footnote{Notice that the decision of considering just the expected value of the pdfs may not be ideal in this case, given the multimodality of these pdfs, as shown in Figure \ref{fig:P_Y_given_Z} in the appendix.}, the head of the department obtains the following fair results:
\[
\begin{array}{ccc}
E\left[P \left(\predY_A\left(\textrm{App}_{1}\right)\vert Z\right)\right]=3.022 & \,\,\,\,\,\, & E\left[P \left(\predY_B\left(\textrm{App}_{1}\right)\vert Z\right)\right]=2.312\\
E\left[P \left(\predY_A\left(\textrm{App}_{2}\right)\vert Z\right)\right]=3.030 &  & E\left[P \left(\predY_B\left(\textrm{App}_{2}\right)\vert Z\right)\right]=2.313
\end{array}
\]
These results are more comforting in that, while they still allow room for disagreement between Alice and Bob over the evaluation of individual candidates, they guarantee that the two applicants, who differ only on a protected attribute, receive identical predictive scores (within the numerical precision of a Monte Carlo simulation\footnote{This precision can be increased by incrementing the number of Monte Carlo samples collected.}).
Again, formally, if we were to see the second candidate as an intervention on the first, we would have:
\[ 
P\left( \predY_{\textnormal{Gnd}\leftarrow F} (\context{u}) \vert X=x, A=a \right) =  
P\left( \predY_{\textnormal{Gnd}\leftarrow M} (\context{u}) \vert X=x, A=a \right),
\]
thus satisfying counterfactual fairness.
Therefore, these counterfactually-fair scores can now be safely averaged into a final fair predictor $\predYpool$ by the head of the department and used for decision-making.

\section{Conclusion \label{sec:Conclusion}}
This paper offers a complete approach to the problem of computing aggregated predictive outcomes from a collection of causal models while respecting a principle of counterfactual fairness. Our solution comprises two phases: (i) a qualitative step, in which we use judgment aggregation to determine a counterfactually-fair pooled model; and, (ii) a quantitative step, in which we use Monte Carlo sampling to evaluate the predictive output of each model by integrating out unfair components, and then we perform opinion pooling to aggregate these outputs. The entire approach was illustrated on the toy-case of PhD admissions, showing that it does indeed provide counterfactually-fair results.

However, this work represents just a first attempt at solving the problem of aggregating multiple causal models in order to provide counterfactually-fair predictions. Some avenues for future development that we are investigating include:

\begin{itemize}
	\item our method presupposes causal models defined over the same set of exogenous and endogenous variables; however, our solution is quite flexible and, with little formal work, it may be extended to produce fair outcomes from the aggregation of causal graphs defined on different sets of exogenous and endogenous variables;
	
	
	\item from a formal point of view, it may be interesting to investigate extreme cases (e.g.: scenarios in which qualitative aggregation provides no fair model), examine what are the conditions for a fair model to exist, and evaluate how these conditions may be relaxed to allow the \emph{most fair possible} aggregation of causal models;
	
	\item more importantly, it may be worth to study how a purely observational approach to fairness may be integrated by a pro-active affirmative approach. In this last more realistic approach, the aim is not only to guarantee unbiased outcomes with respect to the available historical data (which may itself be biased), but purposefully and actively compensate existing bias through policies and interventions.

\end{itemize}

\bibliographystyle{splncs04}
\bibliography{opinionpooling}

\newpage

\appendix
\counterwithin{algorithm}{section}
\counterwithin{figure}{section}

\renewcommand\thesection{Appendix \Alph{section}:}
\renewcommand\thealgorithm{A.\arabic{algorithm}}
\renewcommand\thefigure{B.\arabic{figure}}

\section{Algorithms}

\begin{algorithm}[H]
	\caption{Removal Function}\label{alg:removal}
	\begin{algorithmic}[1]
		\State {\bfseries Input:} $\Nexpert$ graph models $\modeldiagram{\model{j}} = \left( \vertexset{j}, \edgeset{j} \right)$, where the vertex set $\vertexset{j}$ is defined over the exogenous and endogenous variables $\exvarset{} \cup \envarset{}$; a partitioning of the variables $\exvarset{} \cup \envarset{} \setminus \{\predY\}$ into protected attributes $\protectedattribset{}$ and $\featureset{}$.
		\State
		\State Initialize $\removalset{fair} := \exvarset{} \cup \envarset{}$
		\For{$j=1$ {\bfseries to} $\Nexpert$}
		\State $\removalset{\neg} := \lbrace W \vert \left(W \in \protectedattribset{}\right) \vee \left(W \in \descendent{\protectedattribset{}}{\model{j}}\right) \rbrace$
		\State $\removalset{fair} := \removalset{fair} \setminus \removalset{\neg}$
		\EndFor
		\For{$j=1$ {\bfseries to} $\Nexpert$}
		\State Remove from the edge set $\edgeset{j}$ of $\modeldiagram{\model{j}}$ all edges $\edge{V_x}{V_y} \vert \left(V_x \notin \removalset{fair} \vee V_y \notin \removalset{fair} \right)$
		\EndFor
		\State

		\State		
		\Return $\model{j}$
	\end{algorithmic}
\end{algorithm}
\begin{algorithm}[H] 
	\caption{Pooling Function} \label{alg:pooling}
	\begin{algorithmic}[1]
		\State {\bfseries Input:} $\Nexpert$ graphs models $\modeldiagram{\model{j}} = \left( \vertexset{j}, \edgeset{j} \right)$, where the vertex set $\vertexset{j}$ is defined over the exogenous and endogenous variables $\exvarset{} \cup \envarset{}$; a judgment aggregation rule $\jarule{}$.
		\State
		\State Initialize $D$ to the length of the longest path in the models $\model{j}$
		\State Initialize $\Mpool$ by setting up the graph $\modeldiagram{\Mpool}$ in which $\vspool{}=\exvarset{}\cup\envarset{}$ and $\espool{}=\emptyset$

		\For{$j=1$ {\bfseries to} $\Nexpert$}
		\State Initialize the vertex set $\vertexset{j,0}:={\predY}$
		\State Initialize the edge set $\edgeset{j,0}:=\emptyset$
		\EndFor
		
		\For{$j=1$ {\bfseries to} $\Nexpert$}
		\For{$d=1$ {\bfseries to} $D$}
		\State $\edgeset{j,d} := \lbrace \left(\edge{V_x}{V_y}\right) \vert \left(\edge{V_x}{V_y}\right) \in \edgeset{j} \wedge \left( V_x \in \vertexset{j,d-1} \vee V_y \in \vertexset{j,d-1}  \right)  \rbrace$
		
		\State $\vertexset{j,d} := \lbrace V_x \vert \left( \edge{V_x}{\cdot}  \right) \in \edgeset{j,d} \vee  \left(\edge{\cdot}{V_x}\right) \in \edgeset{j,d}  \rbrace$				
		\EndFor
		\EndFor
		
		\For{$j=1$ {\bfseries to} $\Nexpert$}
		\For{$d=1$ {\bfseries to} $D$}
		\State $\forall \left(\edge{V_x}{V_y}\right) \in \edgeset{j,d}$, if $\left( \jarule{\edge{V_x}{V_y}} =1 \right) \vee \left( \espool{} \setunion \{\edge{V_x}{V_y}\} \textnormal{is acyclic} \right)$  then $\espool{} := \espool{} \setunion \{\edge{V_x}{V_y}\}$				
		\EndFor
		\EndFor		
		\State
		\Return $\Mpool$
	\end{algorithmic}
\end{algorithm}

\section{Figures}

\begin{figure}
	\centering
	
	\includegraphics[scale=0.5]{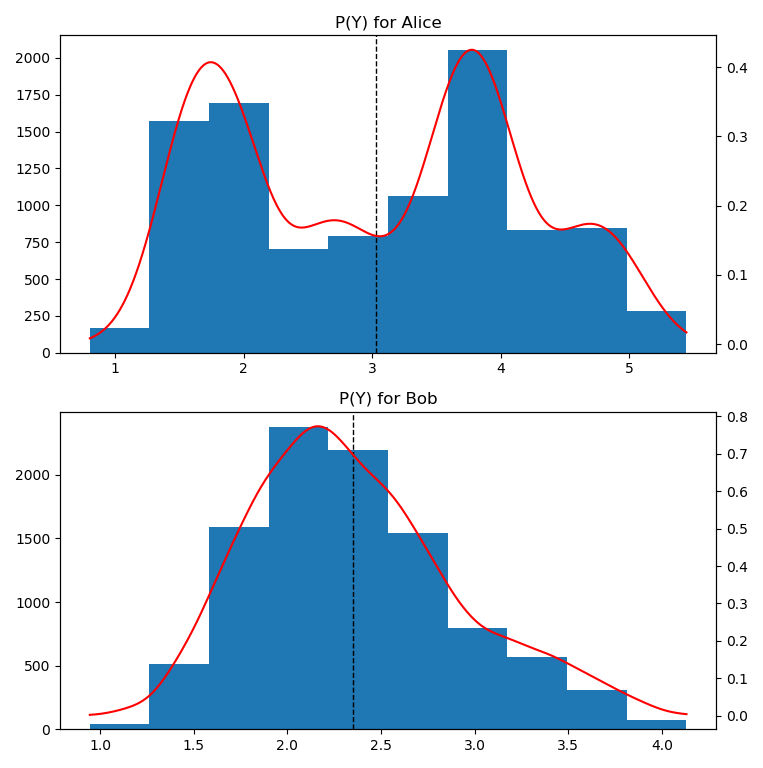}
	
	\caption{Histogram and probability distribution function (computed via kernel density estimation) of $P(\predY)$ in the model provided by Alice and Bob. The \emph{x}-axis reports the domain of the outcome of the predictor $\predY$; the left \emph{y}-axis reports the number of samples used to compute the histogram, while the right \emph{y}-axis reports the normalized values used to compute the pdf. \label{fig:P_Y}}
\end{figure}

\begin{figure}
	\centering
	
	\includegraphics[scale=0.6]{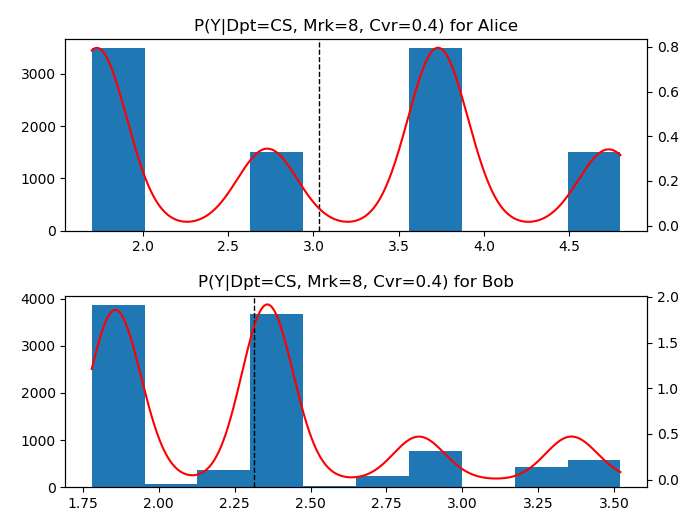}
	
	\caption{Histogram and probability distribution function (computed via kernel density estimation) of $P(\predY \vert  Dpt=\textnormal{CS}, Mrk=0.8, Cvr=0.4)$ in the model provided by Alice and Bob. The \emph{x}-axis reports the domain of the outcome of the predictor $\predY$; the left \emph{y}-axis reports the number of samples used to compute the histogram, while the right \emph{y}-axis reports the normalized values used to compute the pdf. \label{fig:P_Y_given_Z}}
\end{figure}

\end{document}